\documentclass[10pt,twocolumn]{article}

\usepackage{graphicx}
\usepackage{amsmath}
\usepackage{amssymb}
\usepackage{enumitem}
\usepackage{times}
\usepackage{tabularx,booktabs,multirow,makecell}
\usepackage{caption}
\usepackage[linesnumbered,ruled,lined]{algorithm2e}

\newcommand{\x}{\mathbf{x}}
\newcommand{\y}{\mathbf{y}}

\begin{document}


\title{Less is More: Removing Text-regions Improves CLIP Training Efficiency and Robustness}

\author{Liangliang Cao, Bowen Zhang, Chen Chen, Yinfei Yang, \\
Xianzhi Du, Wencong Zhang, Zhiyun Lu, Yantao Zheng \\
Apple AI/ML}

\date{}

\maketitle

\begin{abstract}

The CLIP (Contrastive Language-Image Pre-training) model and its variants are becoming the de facto  backbone in many applications. However, training a CLIP model from hundreds of millions of image-text pairs can be prohibitively expensive. Furthermore, the conventional CLIP model doesn't differentiate between the visual semantics and meaning of text regions embedded in images. This can lead to non-robustness when the text in the embedded region doesn't match the image's visual appearance. In this paper, we discuss two effective approaches to improve the efficiency and robustness of CLIP training: (1) augmenting the training dataset while maintaining the same number of optimization steps, and (2) filtering out samples that contain text regions in the image. By doing so, we significantly improve the  classification and retrieval accuracy on public benchmarks like ImageNet and CoCo. Filtering out images with text regions also protects the model from typographic attacks. To verify this, we build a new dataset named ImageNet with Adversarial Text Regions (ImageNet-Attr). Our filter-based CLIP model demonstrates a top-1 accuracy of 68.78\%, outperforming previous models whose accuracy was all below 50\%.

\end{abstract}

\section{Introduction}

Contrastive Language-Image Pre-training (CLIP) \cite{clip} is a seminal work to build powerful vision-language models with various applications. By learning from billions of image-text pairs, the model performs very well on downstream tasks like zero-shot classification, captioning, retrieval,  segmentation, video recognition, and many others. It has motivated many following works \cite{jia2021scaling} \cite{zhai2022scaling} \cite{yu2022coca}\cite{li2022scaling} \cite{xiao2022exploiting}\cite{flip2023}, which has encouraged the trend of using more training data and bigger models.  

\begin{figure}[t!]
\includegraphics[width=0.5\textwidth]{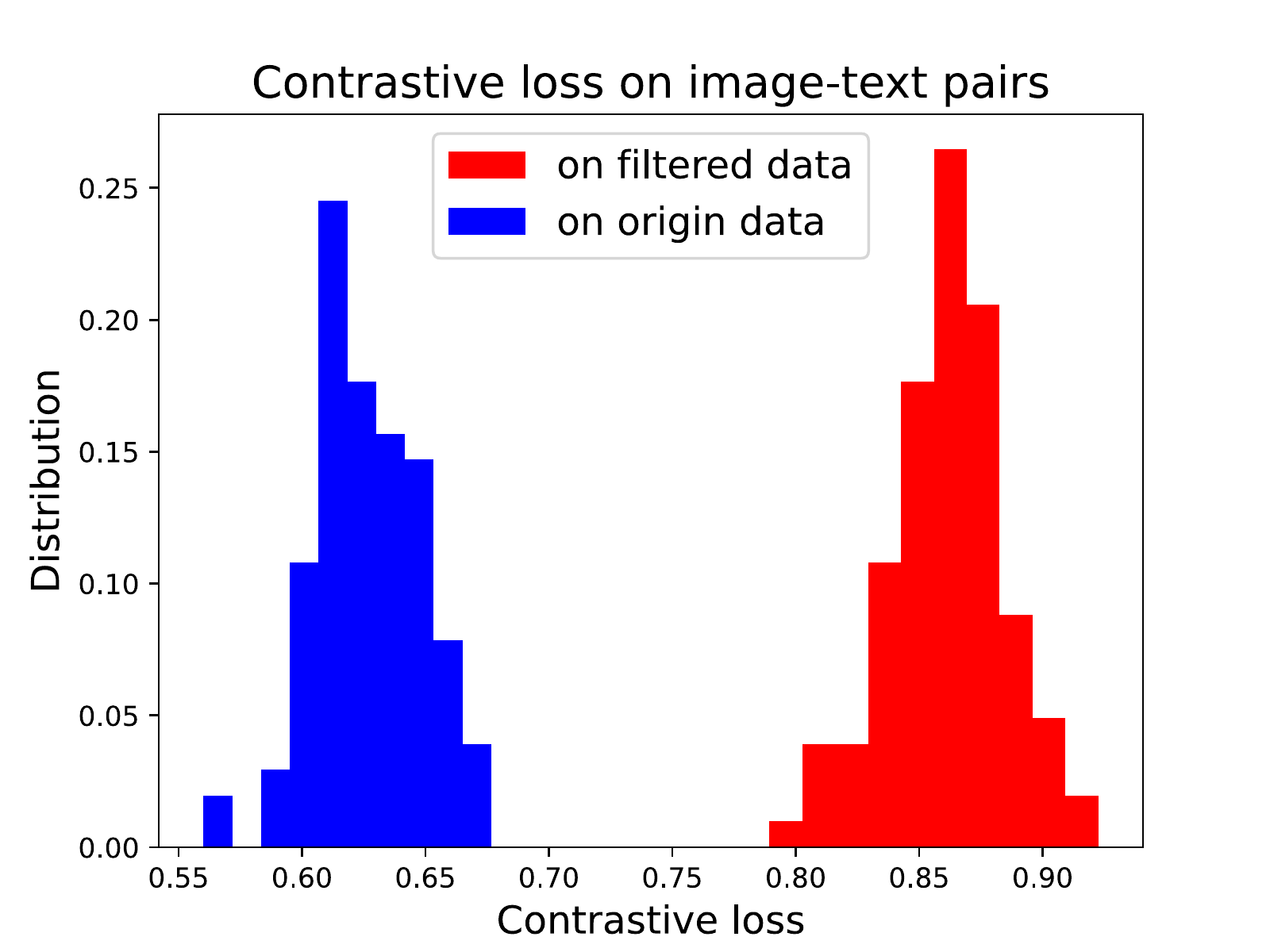}
\caption{The distribution of contrastive loss of CLIP models. Left: contrastive loss on origin image-text pairs. Right: contrastive loss on filtered data where images have no text regions.}
\label{fig-clip-loss}
\end{figure}

\begin{figure}[h!]
\includegraphics[trim=3em 7em 1em 7em,  width=0.45\textwidth]{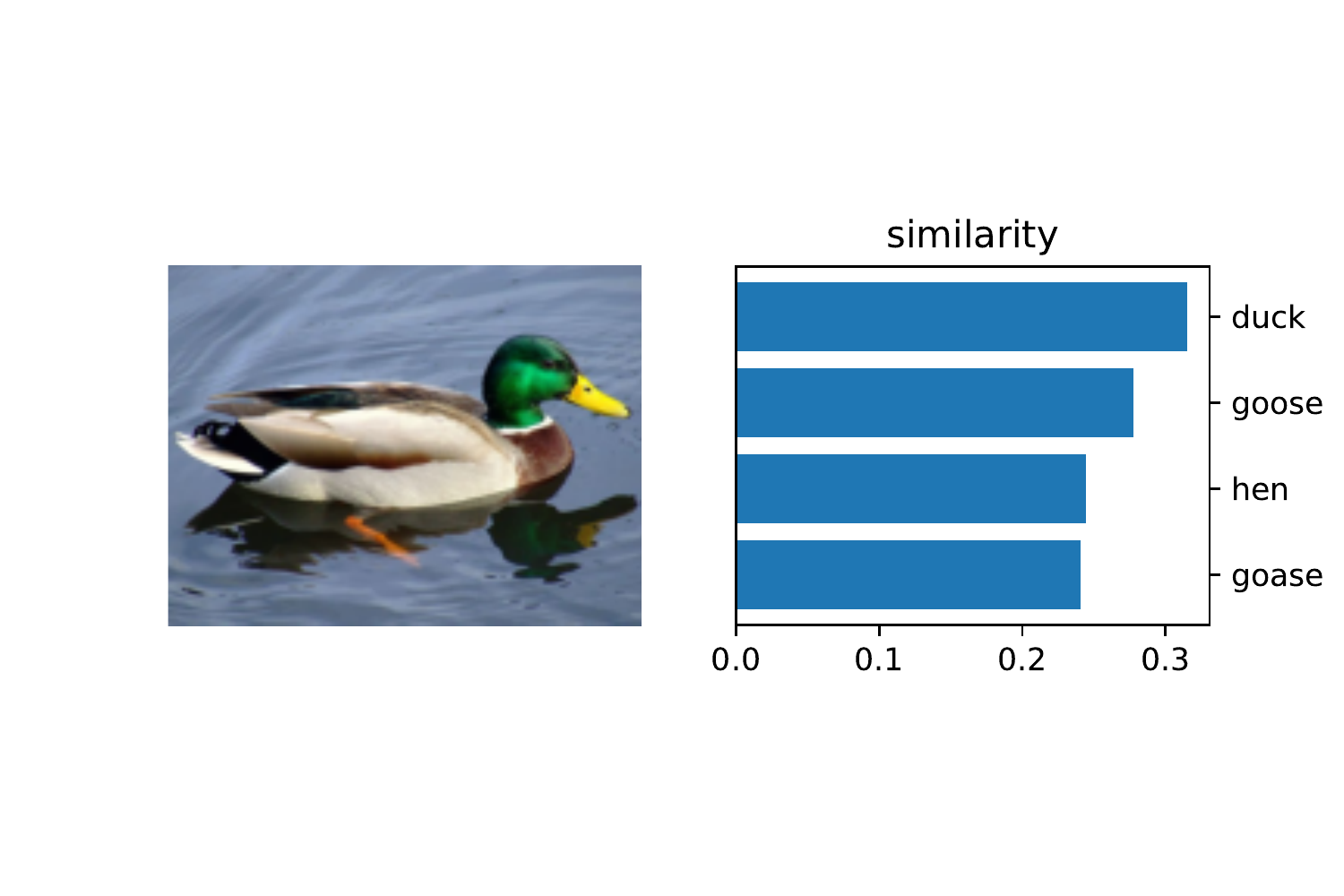}
\includegraphics[trim=3em 7em 1em 7em, width=0.45\textwidth]{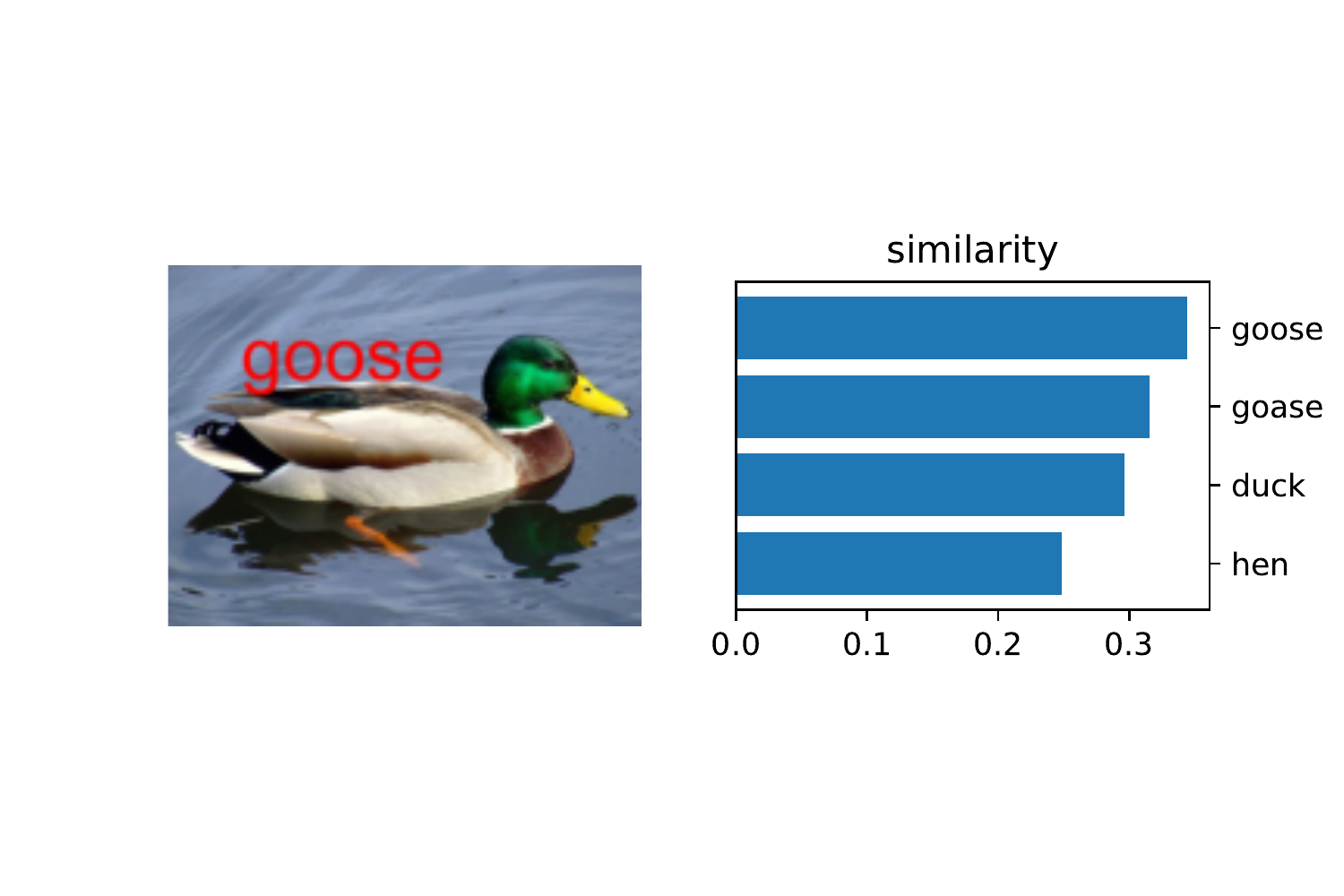}
\caption{The similarities from CLIP image-text embeddings are misled by the text regions whose meaning is inconsistent with the visual semantics.}
\label{fig-text-region-attack}
\end{figure}

Since training CLIP is expensive, this work considers the scenario with a fixed optimization budget and discusses a few general but simple-to-use techniques to improve the CLIP models' training efficiency and robustness. Our study is motivated by the observation of contrastive loss in Fig.~\ref{fig-clip-loss}, where we find the distribution of contrastive loss will change dramatically if we only consider these image pairs without text region. This suggests that the traditional CLIP models match text semantics better than visual semantics. Fig.~\ref{fig-clip-loss} implies that if we focus on visual semantics (red bars), we may improve the CLIP training more efficiently.

In practice,  the semantics of text regions may or may not match visual semantics. Figure~\ref{fig-text-region-attack} shows a failed example of CLIP-based zero-shot recognition. Although the original image can be recognized correctly as ``duck", the model will fail if we add a text region of ``goose". Interestingly, with the text region, the model may get confused with ``goase" which has similar text tokens but no correct meaning.   
Such images are also called ``typographic attacks" \cite{goh2021multimodal}. The failure in Fig.\ref{fig-text-region-attack} suggests the non-robustness of CLIP models, which may hurt the performance in recognition and retrieval tasks when the text does not match visual semantics.

In this paper, we want to kill two birds with one stone. We fix the computational budget (i.e., same number of optimization steps and batch sizes) and explore how to improve the performance of CLIP models. We found two seemingly contradictory approaches to improving training accuracy within a fixed training budget: On one hand, incorporating more training examples leads to lower training loss when maintaining the same training budget (i.e., fewer epochs). On the other hand, pruning the training set without text regions can further boost the efficiency and stability of the model. In addition, our experiments show that filtering data with text regions will 
force the model to focus on image content instead of text regions and thus avoid the mistakes in Figure~\ref{fig-text-region-attack}. In this way, we improve both the efficiency and robustness of the CLIP models.

One potential limitation of our work is that the CLIP model trained in this paper may lose the ability to understand embedded texts (i.e., optical character recognition). We argue that OCR is fundamentally a different problem from visual understanding and should be solved by a separate module. In addition, OCR modules usually use a small network \cite{ocr-gooogle-cvpr2022} for irregular-shaped text regions \cite{ocr-binary-detection-pami22}, so that in practice, we propose to treat OCR as a different task than general visual understanding.

The contribution of this paper is three-fold: (1) We compare different ways of improving CLIP training data and recommend a simple approach of filtering out data with text regions. (2) We build a new evaluation dataset to benchmark the robustness against typographic attacks. (3) Extensive experiments demonstrated the filtering approach consistently outperforms the baselines by improving the top-1 accuracy on ImageNet from 68.66\% to 70.77\%, and more significantly, on our new evaluation set, from 35.73\% to 68.78\%. 

\section{Related Works}

Quite a few works have discussed how to improve the training data for CLIP-like models.   ALIGN \cite{jia2021scaling}  has discussed many filtering tricks for selecting the training data. BASIC \cite{pham2021combined} scaled both data size and batch size with a larger backbone model. More recently, LAION \cite{laion2b} collected a large open-sourced dataset and trained a large G/14 model on the 2B dataset. \cite{zhai2022scaling} introduces a  gigantic model with 2B parameters, and the corresponding models are usually too expensive to be deployed to large-scale production. Similarly, \cite{declip2021} compares different supervision signals to train CLIP-like models more efficiently. \cite{radenovic2023filtering} discusses various techniques on filtering, distillation, and hard negative mining for CLIP pre-training. \cite{yao2021filip}  extends CLIP to fine-grained scenarios. FLIP \cite{flip2023} explores image masking to improve pre-training. Some recent works \cite{xCLIP2022} and \cite{zhai2023sigmoid} discuss ways to do the pre-training with non-contrastive losses. However, most of these works are based on heuristic insights rather than rigorous analysis. Their consensus is to assemble a bigger dataset with a bigger model. Limited guidance on improving CLIP with a fixed optimization budget exists.

In spirit,  our work is partially motivated by  Chinchilla \cite{Chinchilla2022}, a classic work in large language modeling. Chinchilla \cite{Chinchilla2022} used the same compute budget (FLOPs) as Gopher \cite{Gopher2022} but with a smaller number of parameters and four times more data, but outperforms the baseline on many NLU benchmarks. 
However, Chinchilla is devoted to large language models but not training with image data.  
In this paper, we show that beyond increasing the size of the training data, sometimes it is also useful to reduce the training set to help the performance of CLIP training. Our conclusion implicitly suggests that vision-language models are still different from large language pre-training and encourage more studies in the data pruning \cite{sorscher2022beyond} \cite{lu2022unsupervised} direction.

Other research studies are also inspiring to us. The first group includes the new algorithms to train a CLIP-like model more efficiently \cite{li2022scaling} \cite{reclip2023} \cite{chen2023symbolic} \cite{zhai2023sigmoid}. For the sake of simplicity, this paper chooses the standard CLIP-B/16 and CLIP-L/14 models for experiments. The conclusion of our studies may also apply to not these modified models, and we will leave it for future studies. The second group includes the application of using CLIP models for OCR tasks. Starting from the pioneering work \cite{clip}, many works \cite{yang2021tap} \cite{kil2022prestu} \cite{song2022vision} \cite{li2021trocr} \cite{xue2022languagematters} show that we can borrow or extend the CLIP model to recognize the texts in the images. The purpose of this paper is think in a reverse direction; we find it is beneficial to deprive the OCR capability of CLIP model to gain more efficiency and robustness of the image content understanding. We will delve into this topic in detail in later sections.

\section{Improving Data for CLIP Training}

\subsection{Primary Model}
In this paper, we study the vanilla CLIP model. Our dataset follows the collection discussed in \cite{chen2023stair}, which is a combination of internal and public datasets. The public datasets consists of Conceptual Caption 3M (CC-3M)~\cite{sharma-etal-2018-conceptual} and Conceptual Captions 12M (CC-12M)~\cite{changpinyo2021cc12m}.
The internal image-text dataset consists of 1B image-text pairs, including a 134M clean licensed dataset and a 971M noisy web-crawled dataset. The web-crawled dataset is mined following the approach described in ALIGN~\cite{jia2021scaling} and CLIP~\cite{clip}. Note that due to license constraints, we cannot use the Laion dataset, but the performance of our baseline is comparable with the B16 model reported in the original CLIP paper \cite{clip} as well as the CLIP B16 trained on Laion-400M\cite{laion2b}.

We follow CLIP-B/16~\cite{clip} as our primary model. The text encoder is a 12-layer transformer \cite{transformer} with 512 hidden dimensions and 8 attention heads. The text input is tokenized by BERT WordPiece tokenizer~\cite{kenton2019bert}  with 30,522 vocabularies. The max input sequence length is set to 76. The image encoder is a 12-layer visual transformer \cite{vit2020} with 12 attention heads and 768 hidden dimension sizes. We implement the CLIP model using Jax and train it with 256 TPUs.  Note that CLIP training requires a large batch size, and we find enlarging the batch size to 32K obtains better performance than 16K, while comparable with the batch size of 64K. The baseline CLIP model is trained with 340K steps using an AdamW optimizer with a learning rate of $5e^{-4}$  and a weight decay ratio of $0.2$. The learning rate is first warmed up till 2000 steps, and then cosine decay to zero. During the optimization, each batch includes 32K pairs of images and texts.
 
\subsection{Comparing Different Training sets}

\begin{figure}[h!]
\includegraphics[width=0.48\textwidth]{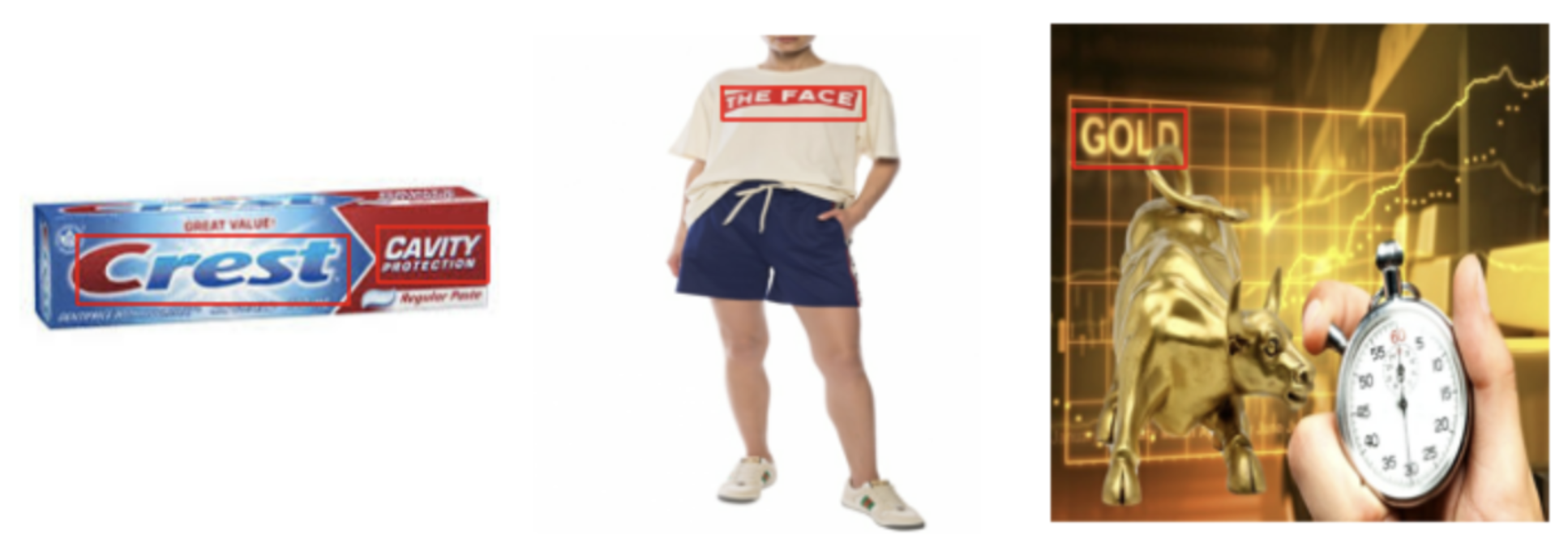}
\caption{Example of training data with text regions. The text regions are marked with red bounding boxes.}
\label{fig-example-text-regions}
\end{figure}

By observing our training data, we found about 40\% of the training images include text regions. Fig.~\ref{fig-example-text-regions} illustrates some examples. Note that images with text regions are very popular from the web, which is the major reason why the ratio of such regions is high. 

We try different approaches to improve CLIP models and compare them with the baseline. To use the same optimization budget as the baseline, we fix the CLIP models and optimization budget (i.e., 340K steps with the same batch size), but with different training data:

\begin{description}[leftmargin =*] 
    \item[Origin-1.1B] The original dataset with 1.1B image-text pairs. 
    \item[Origin-0.7B] Sample source as Origin-1.B, but randomly sampled 0.7B pairs. 
    \item[Filter-0.7B] Filtering those pairs with text regions in the images, which filters out 40\% of the data in Origin-1.1B, and leaves 0.7B image-text pairs.  
    \item[Blur-1.1B] Blurring to the text regions in the images, which leads to a training set with 1.1B pairs. 
\end{description}

For our implementation, we have utilized the CRAFT detector \cite{ocr-craft-cvpr2019} to determine the presence of text regions within an image. However, we acknowledge that many other OCR libraries are available that may provide similar or superior results. For Blur-1.1B, we first detect the text regions and then apply a Gaussian blur to ensure that the text is unreadable by humans. In our workflow, we first resize the image to $224\times 224$ and then apply a Gaussian blur with a radius of $15$.

We first examine their training loss to compare  CLIP models trained from different datasets. Given a batch of image-text pairs $\x_i, \y_i$, with $1\leq i \leq |B|$, the contrastive loss over the image-text pairs, which is widely used for CLIP training:
\begin{align}
l_c 
=&-\frac{1}{2|B|} \sum_i (\log\frac{\exp(\x_i\cdot\y_i/T)}{\sum_{j=1}^{|B|} \exp(\x_i\cdot\y_j/T)} + \nonumber \\
&+\log\frac{\exp(\x_i\cdot\y_i/T)}{\sum_{j=1}^{|B|} \exp(\x_j\cdot\y_i/T)})
\label{eq-clip-loss}
\end{align}
where $\x_i$ and $\y_i$ are normalized embedding vectors for images and texts. 
$T$ is a temperature parameter to normalize the softmax function. In practice, we follow  \cite{clip} to use the logit scale to clip the temperature. This paper considers contrastive losses to compare models trained after the same number of steps with the same temperature. Based on our computational budget, we limit the learning steps to 340k steps and fix the batch size as 32k pairs, which translates to 11 billion samples\footnote{Our computational budget is comparable with that used by OpenAI CLIP and LAION CLIP B16.}.  

\begin{figure*}[t!]
\includegraphics[width=0.45\textwidth]{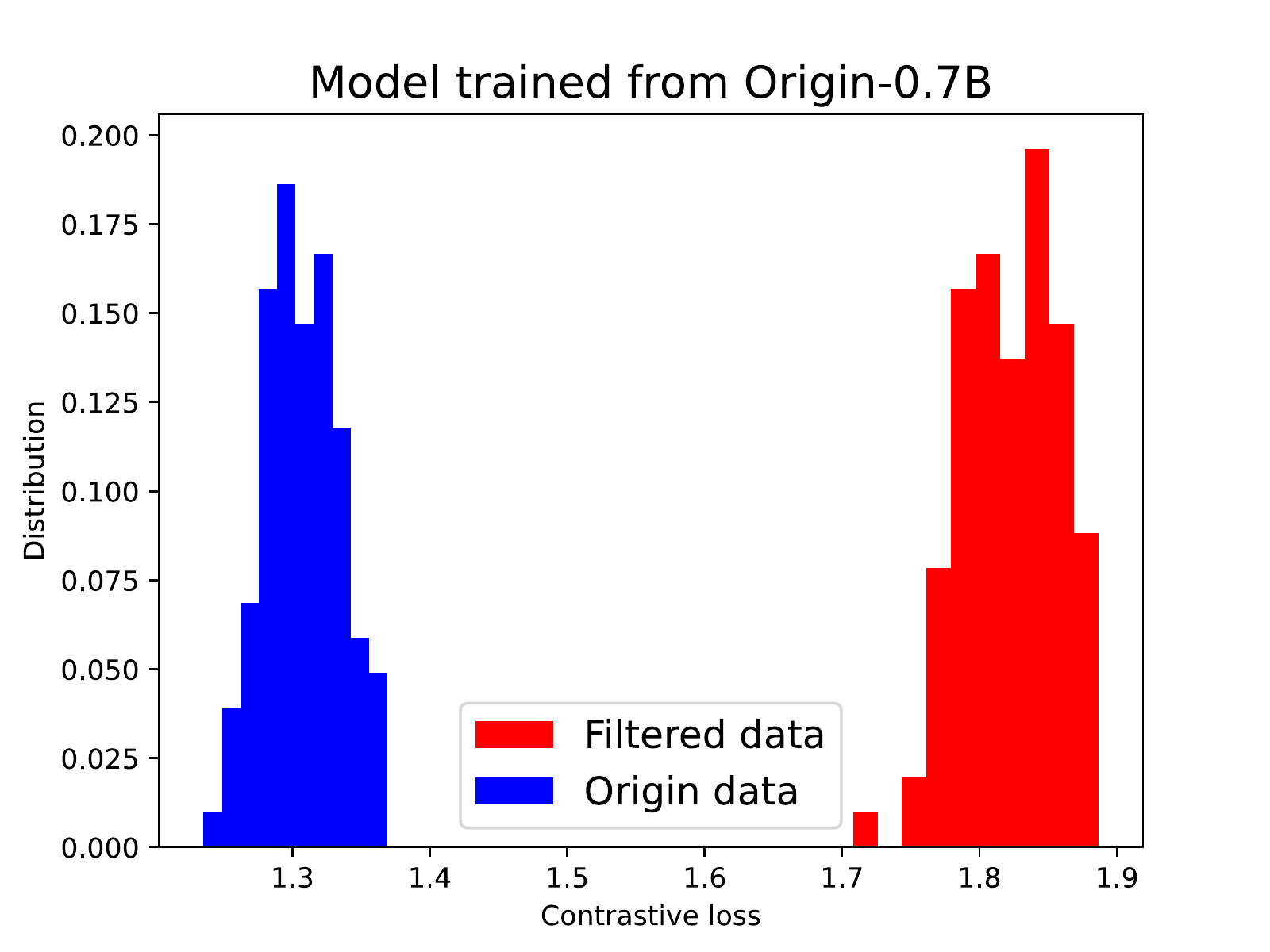}%
\includegraphics[width=0.45\textwidth]{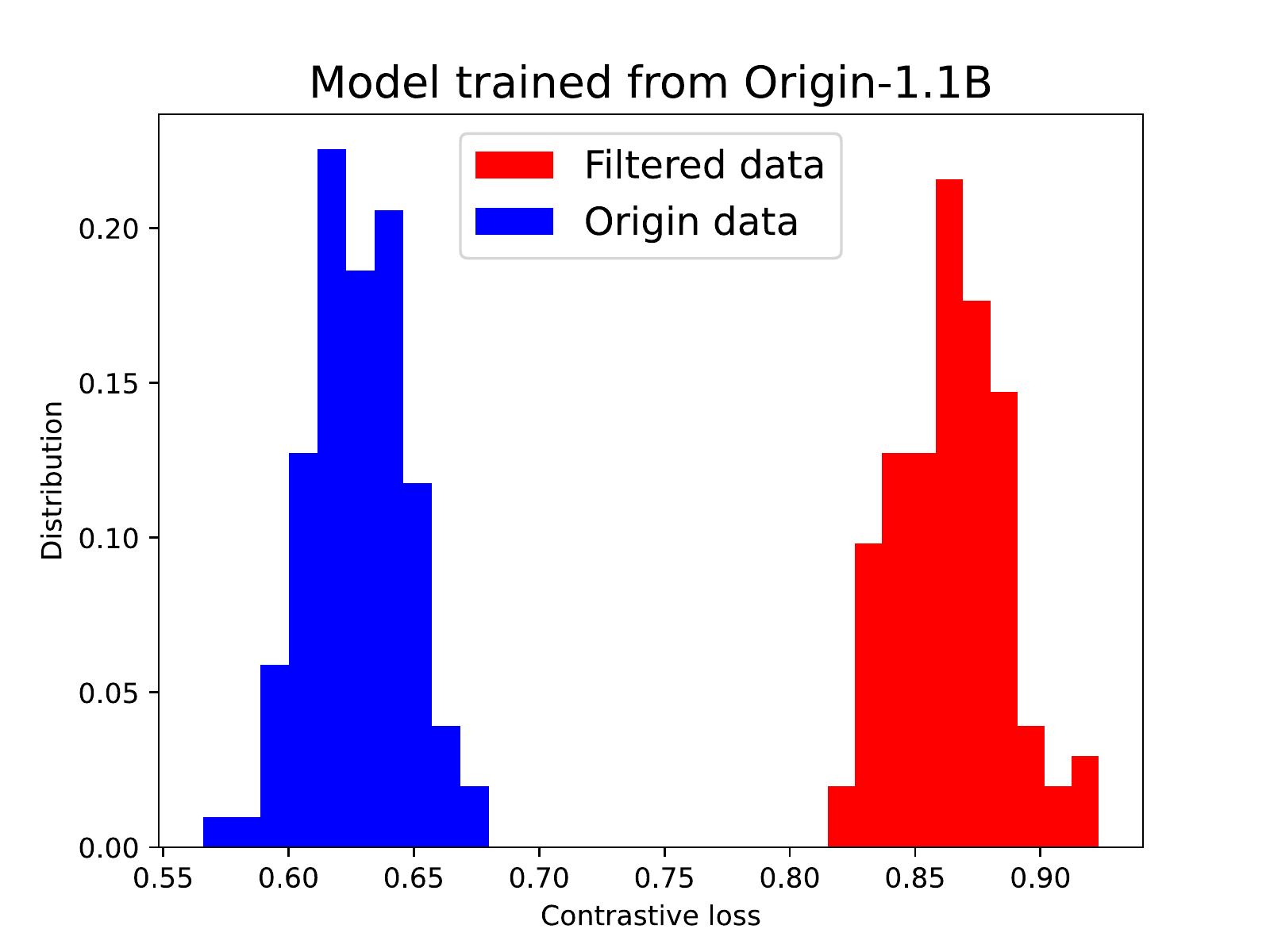} \\
\includegraphics[width=0.45\textwidth]{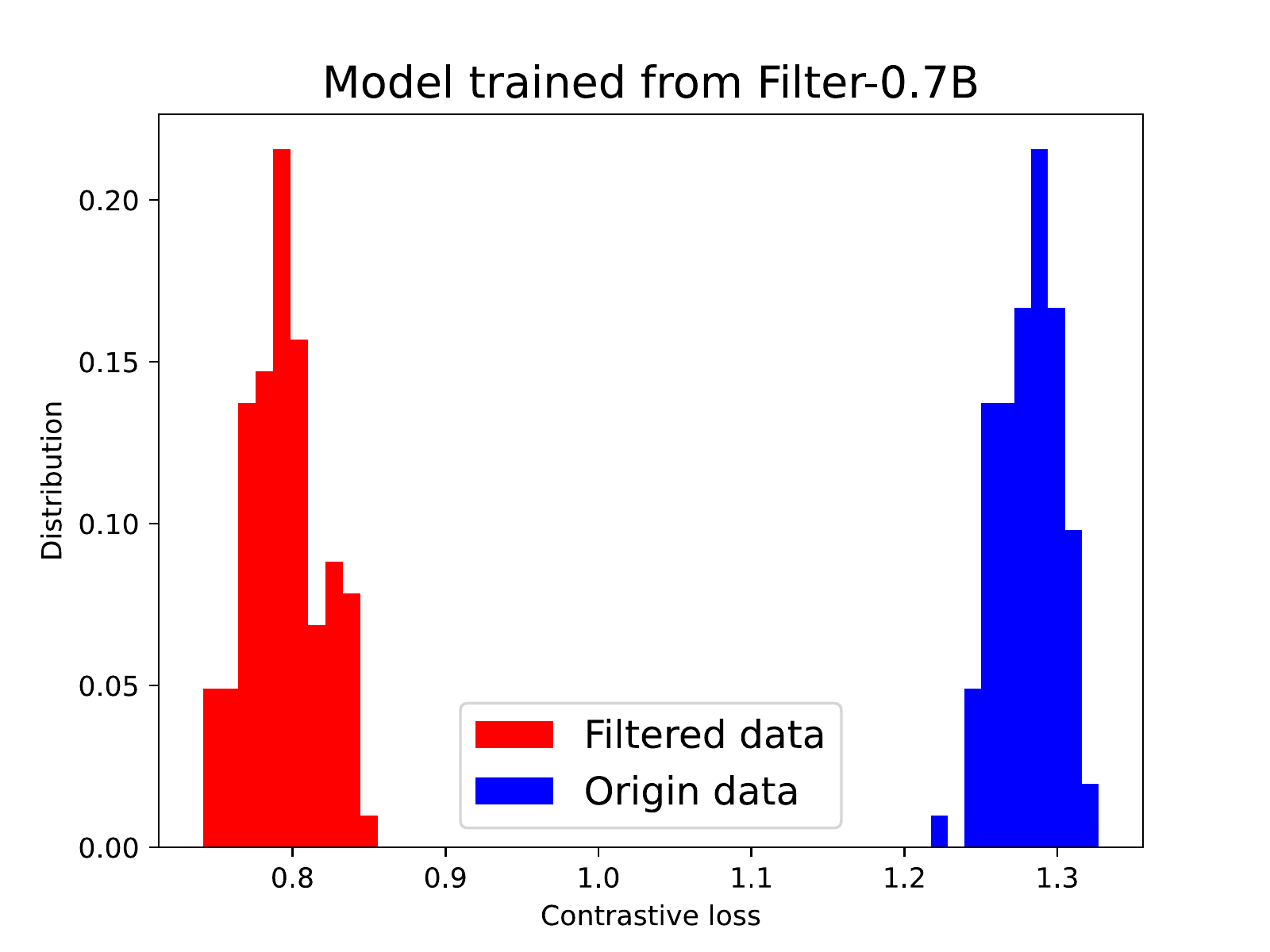}%
\includegraphics[width=0.45\textwidth]{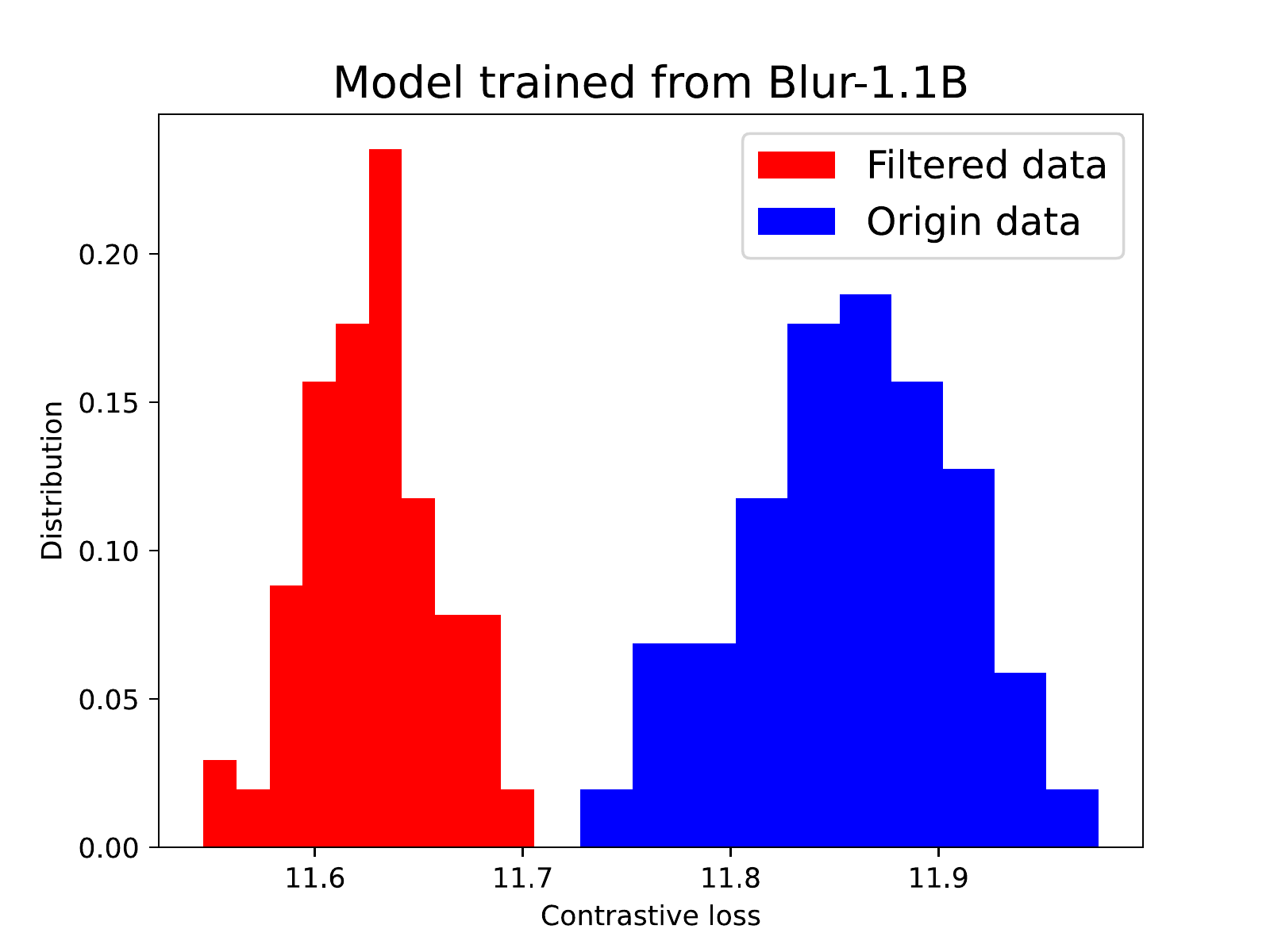} 
\caption{Compare the distribution of contrastive loss  trained from Origin-0.7B, Origin-1.1B, Filter-0.7B and Blur-1.1B. Blur bars correspond to the loss on the origin dataset, while red bars correspond to the loss on the filtered dataset. We can see that on Origin-0.7B and Origin-1.1B, the blue bars' scores are lower than those of red bars. In contrast, on Filter-0.7B and Blur-1.1B, the scores of blue bars are higher than red bars. }
\label{fig-clip-loss-distri}
\end{figure*}

Figure~\ref{fig-clip-loss-distri} compares the contrastive losses from different models. We random sample 100 batches from the original dataset and plot the distribution of corresponding losses in blue. In addition, we sample another 100 batches and plot the distribution in red. We can see that for models trained from Origin-1.1B and Origin-0.7B, the losses from origin batches (blue bars) are significantly lower than those from filtered batches without text regions (blue bars). However, for models using Filter-0.7B and Blur-1.1B, losses from origin batches (blue bars) are higher than those without text regions (red bars). That suggested when we choose Filter-0.7B or Blur-1.1B, the CLIP model will focus on data without text regions.

It will be interesting to compare the models trained from Filter-0.7B and Blur-1.1B quantitatively. Table~\ref{tab-comparing-contrastive-loss} compares the mean and standard deviation of contrastive losses on the 100 batches without text regions. We can see that although Origin-1.1B and Blur-1.1B have more training examples, models trained from Filter-0.7 get the lowest loss values.

\begin{table}[h!]
  \caption{ Compare the contrastive loss on patches without text regions.} 
  \centering
  \begin{tabular}{l|cc}
    \toprule
   Model & Training Steps & Contrastive loss \\
    \midrule
    Origin-0.7B &340K  & $1.8114\pm 0.0405$ \\
    Origin-1.1B &340K & $0.8662\pm 0.0202$\\
    Filter-0.7B &340K & \textbf{$0.8029 \pm  0.0228$} \\
    Blur-1.1B   &340K  &  $0.8823\pm 0.0251$ \\
    \bottomrule
  \end{tabular}
  \label{tab-comparing-contrastive-loss}
\end{table}

 \noindent\textbf{Zero-shot classification and retrieval}: Since CLIP is trained with a massive amount of data, it is good at a wide range of tasks including zero-shot classification and retrieval. For zero-shot classification, we can take the category names of different classes as the set of potential text pairs and predict the most possible (image, text) pair according to CLIP. In practice, the category names work better with certain prompts like "a photo of" and "an image of". We borrow  80 prompts from \cite{clip} and compute 80  embeddings of text prompt + a category name using the CLIP text encoder. In practice, we apply L2-normalized to embedding vectors and then calculate their inner products. Similarly, we can compute the embedding for every image using the CLIP image encoder. To find the most similar class, we compute the cosine distances between image embedding and the averaged text embedding vector, and find the category name which maximizes the cosine distance. Similarly, we can apply CLIP to retrieval tasks, including text-to-image (t2i) retrieval and image-to-text (i2t) retrieval. Following the previous work, we use ImageNet2012 to compare the zero-shot classification task and CoCo to compare the t2i and i2t retrieval tasks. 
Table~\ref{tab-comparing-zeroshot-retrieval} suggested that the model trained from Filter-0.7B outperforms all the other variants.   When using the base model (CLIP B-16), the zero-shot top-1 classification on ImageNet improves from 67.34\% to 68.18\% after we enlarge the training set from 330M image-text pairs to 700M pairs. After we enlarge the training set to 1.1B pairs, we obtain 0.6866. More interestingly, if we decompose the training set into two disjoint sets, e.g., 410M images with text regions and 690M images without text regions, and keep only the latter, we find the model trained from 690M pairs can obtain a higher accuracy of 0.7077. We observed similar trends on CoCo retrieval benchmarks.

\begin{table}[h!]
  \caption{ Comparing zero-shot classification and retrieval tasks.} 
  \centering
  \begin{tabular}{l|cccc}
    \toprule
   Training data & ImageNet Top-1 Acc & Coco i2t &  CoCo t2i \\
    \midrule
    Origin-0.7B  &68.18\% & 57.32\%& 41.31\% \\
    Origin-1.1B  &68.66\% & 57.36\%& 41.34\%\\
    Filter-0.7B &\textbf{70.77\%} &\textbf{58.30\%} &\textbf{42.54\%} \\
    Blur-1.1B & 68.34\% & 57.84\% & 41.48\%\\
    \bottomrule
  \end{tabular}
  \label{tab-comparing-zeroshot-retrieval}
\end{table}

 \noindent\textbf{Finetuning Tasks}:
We also explore which model provides the best image presentation for downstream tasks. Following \cite{clip} \cite{chen2023stair}, we compare the linear probe performance on ImageNet. For the training images from ImageNet data, we compute the embedding vectors using the visual encoder of different CLIP models and train a linear classifier. 
Table~\ref{tab-comparing-linear-prob} compares the performance of different models. The model trained from Filter-0.7B outperforms the other models, suggesting that its visual feature presentation may be attractive for downstream applications. 

\begin{table}[h!]
  \caption{ Linear probing accuracy on ImageNet.} 
  \centering
  \begin{tabular}{l|c}
    \toprule
   Model & Linear Probe Accuracy \\
    \midrule
    Origin-0.7B  & 79.67\% \\
    Origin-1.1B  & 80.29\% \\
    Filter-0.7B & \textbf{80.56\%} \\
    Blur-1.1B &  79.75\% \\
    \bottomrule
  \end{tabular}
  \label{tab-comparing-linear-prob}
\end{table}

\subsection{Analysis}

As discussed in  \cite{bottou2010-sgd}, when we fix with the same amount of optimization time (i.e., learning steps), the learning error is bounded with 
\begin{align}\label{eq-error-bottou2010}
\xi =&\xi_{app} + \xi_{est} + \xi_{opt} \\
\sim&\xi_{app} + (\frac{\log n}{n})^\alpha + \rho \\
&\mathrm{~for~some~}\alpha\in[\frac{1}{2}, 1] \nonumber
\end{align}
where  $n$ stands for the number of training data,  $\rho$ is a pre-defined tolerance for optimization, and the approximation error,  $\xi_{app}$ measures how closely optimal solution $f^*$ can be approximated by a chosen family of functions defined by network $\mathcal{F}$.

\noindent\textbf{More training samples}: From eq~\eqref{eq-error-bottou2010}, we can see that
 the error rate will decrease with $(\frac{\log n}{n})^\alpha$.  So the error $\xi$ will decrease with larger $n$. In other words, if we enlarge the size of the training set while still keeping the same amount of training steps, we will get a better model with lower errors.

\noindent\textbf{More focused training}: When the network structure and $n$ is fixed, $\xi_{app}$ and $\xi_{est}$ will not changed. Thus we can focus on 
 $$
  \xi_{opt} = \mathbb{E}[E(\hat{f}_n) - E(f_n)]
 $$ 
 where $E(f_n)$ stands for the empirical loss with $n$ examples, while
 $E(\hat{f}_n)$ corresponds to the CLIP's contrastive loss during the optimization. When we filter out images with text regions, the model can be more focused and obtain smaller optimization errors.

\begin{figure}[t!]
\includegraphics[width=0.5\textwidth]{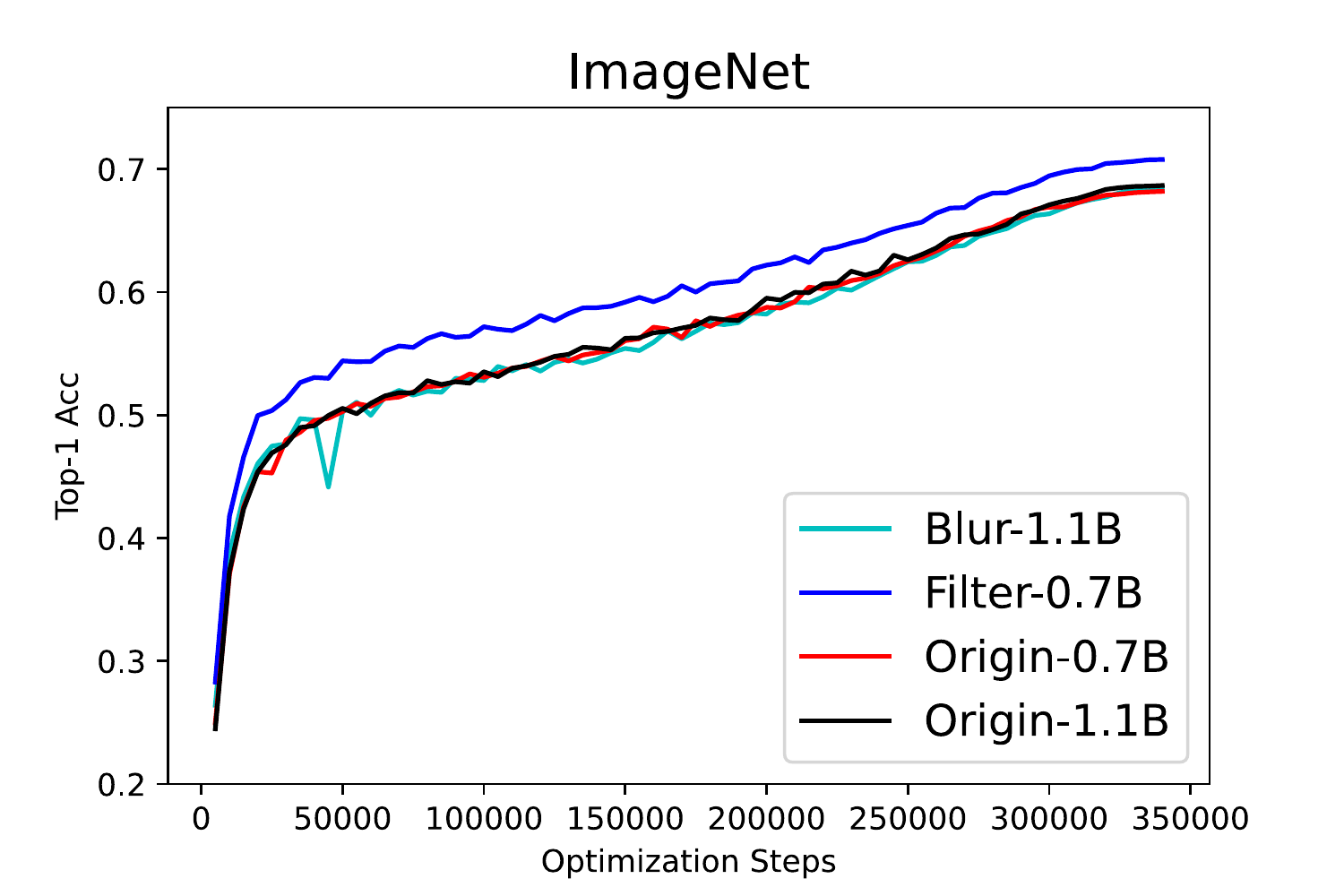}
\caption{ImageNet accuracy from different models.}
\label{fig-clip-optimization-imagenet}
\end{figure}

The studies presented in this section propose two seemingly opposing strategies to enhance training accuracy while keeping the training budget constant: (I) Including a larger number of training examples leads to lower training loss when the same training budget is maintained (i.e., with fewer epochs). (II) Excluding images containing text regions can also improve the model's efficacy and robustness. As Table~\ref{tab-comparing-zeroshot-retrieval} shows, when we enlarge the origin training set from Origin-0.7B to Origin-1.1B, the accuracy on ImageNet and CoCo improves. Furthermore, if we filter out images with text regions, the model from Filter-0.7B significantly outperforms the other approaches.

\begin{figure}[t!]
\includegraphics[width=0.45\textwidth]{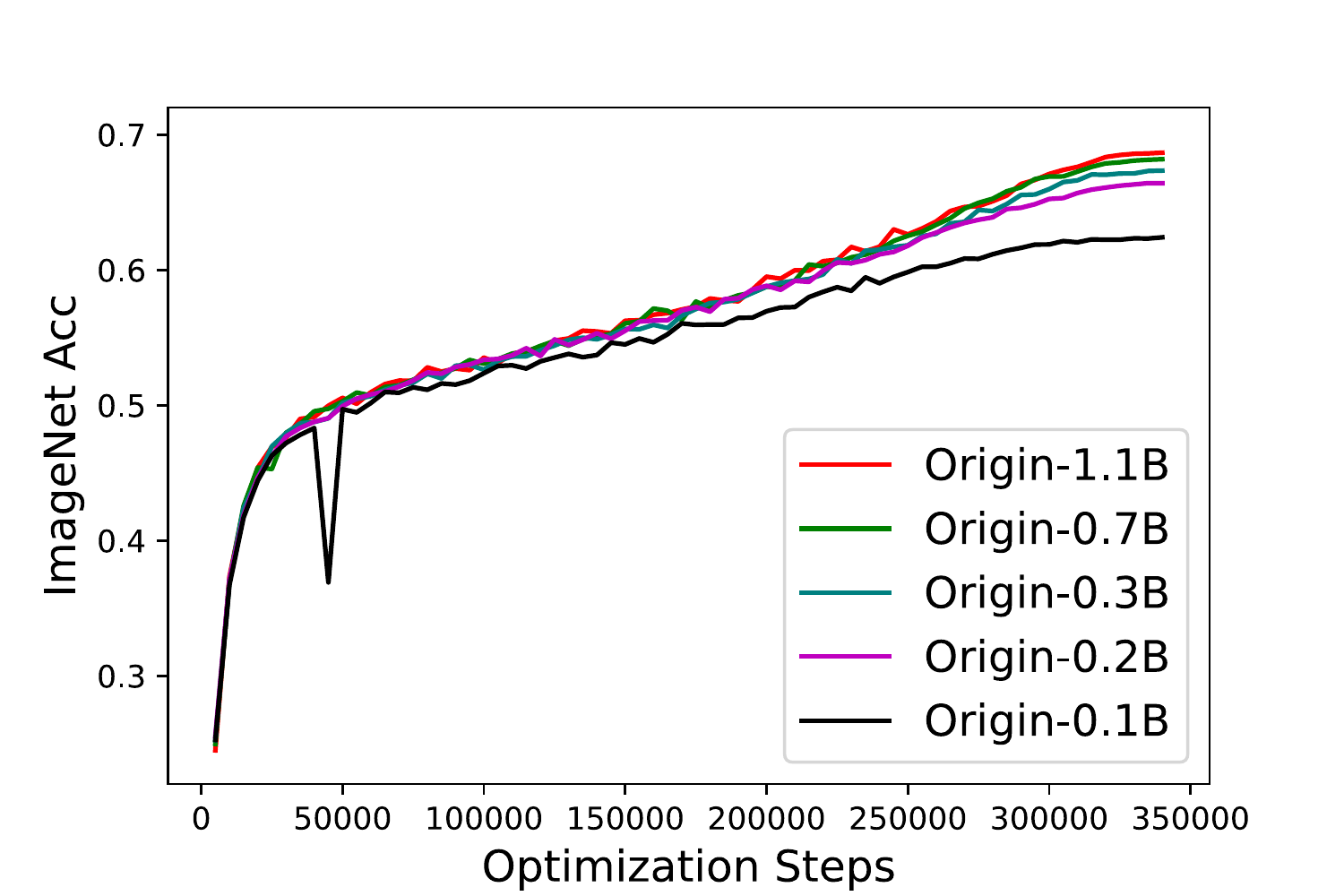}
\caption{Comparing CLIP models by sampling from origin dataset with different sampling ratio.}
\label{fig-sample-vs-acc}
\end{figure}

Figure~\ref{fig-clip-optimization-imagenet} compares
the ImageNet accuracy during the whole training stage. We can see that the model trained from Filter-0.7B significantly outperforms the other approaches throughout the training stage with a good margin. This suggests that combining approach (I) and approach (II) are very effective. To further explore this, Figure~\ref{fig-sample-vs-acc} considers more sampled versions from the origin training data. From Figure~\ref{fig-clip-optimization-imagenet} and Figure~\ref{fig-sample-vs-acc}, we can see Origin-1.1B is significantly better than Origin-0.1B and Origin-0.2B, while Filter-0.7B is significantly better than Origin-1.1B. 

\section{Evaluating against typographic attacks}

\subsection{A New Evaluation Set}

The example shown in Figure~\ref{fig-text-region-attack} shows that the CLIP model will suffer from typographic attacks. In practice, the classic CLIP model will fail when the image contains text regions whose meaning differs from the visual semantics. We want to test if the model trained from Filter-0.7B and Blur-1.1B may suffer less from this problem.

We build a new evaluation set by adding spotting words to the images of ImageNet evaluation sets. There are 1,000 categories in ImageNet. For each category $c$, we find its most confusing category $c'$ and spot the category name to every evaluation image. 

\begin{figure*}[t!]
  \centering
  \includegraphics[height=3cm]{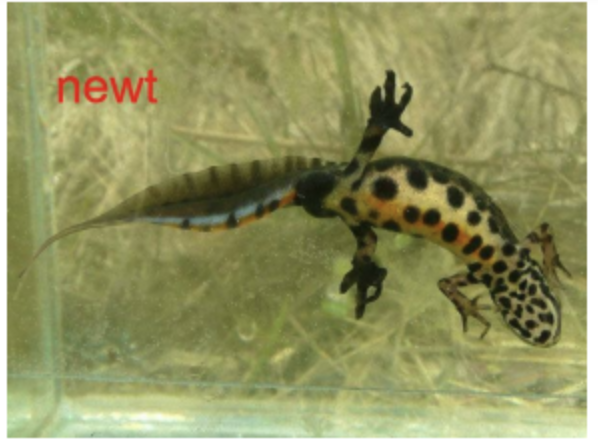}%
  \includegraphics[height=3cm]{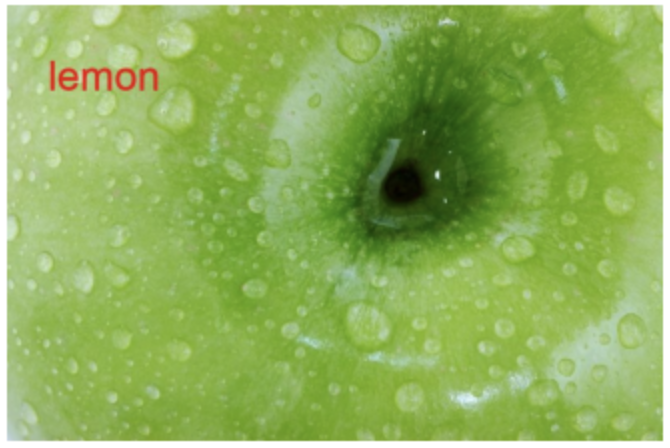}%
  \includegraphics[height=3cm]{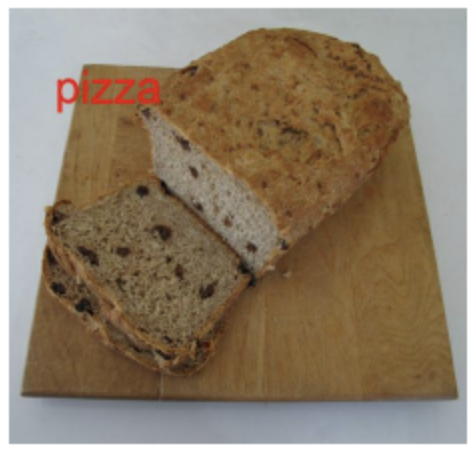}
  \includegraphics[height=3cm]{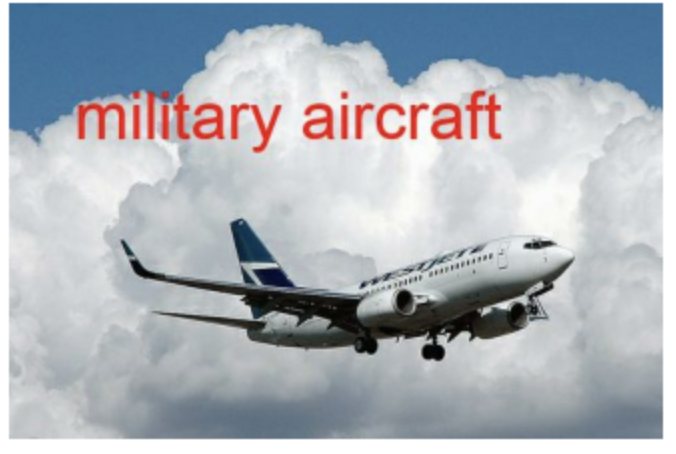}%
  \includegraphics[height=3cm]{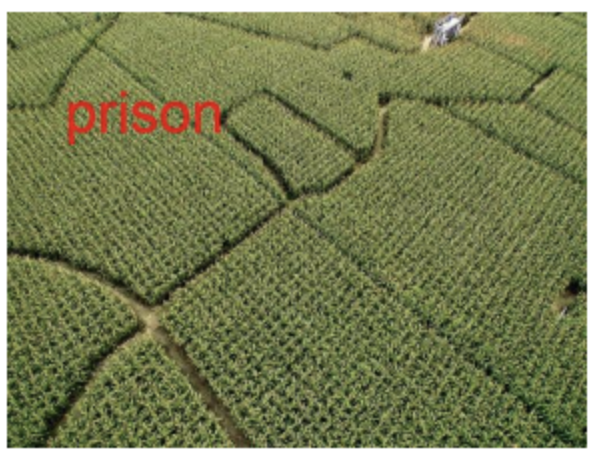}%
  \includegraphics[height=3cm]{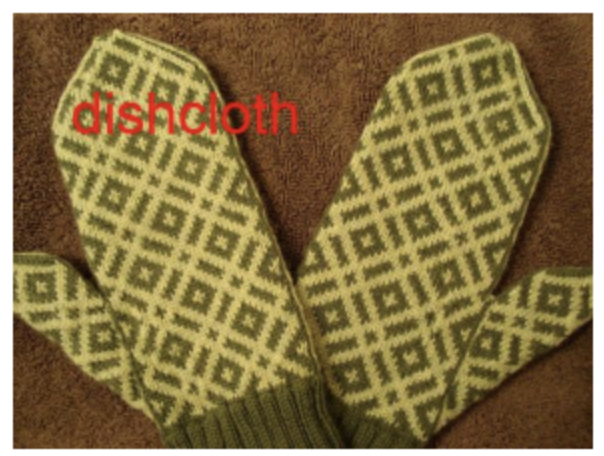}
  \caption{Examples of new ImageNet-Atr dataset. The images are the same as those from ImageNet2012 evaluation set, but we add the text from a confusing category to every image. }
 \label{fig-imagenet-atr-example}
\end{figure*}

To minimize the overfitting problem, we do not use the model trained from our data but the open-sourced OpenAI B/16 CLIP model to compute the confusion matrix. We only choose 1\% of the eval set to calculate the confusing category. However, if all the samples in a category are correctly recognized, we cannot find the most confusing category. In this case, we will use text embeddings $P(w_c)$ to find the most confusing category: 
\begin{equation}
        c^*= \left\{
        \begin{array}{ll}
            \arg \max_{c' \neq c} C(c,c') &\text{If~}C(c,c')>0\\
            \arg\max_{c' \neq c} P(w_c)^T \cdot P(w_{c'}) &\text{Otherwise}
        \end{array}
        \right.
\label{eq-most-confusing-category}        
\end{equation}
where $w_c$ denotes  the  name of category $c$ and $P(w_c)$ denotes the text embedding vector. In our implementation, we borrow the 80 text prompts provided by the origin CLIP paper \cite{clip}, calculate the average of 80 vectors, and then normalize the embedding vector, i.e., $||P(w_c)||^2=1$. 

Algorithm~1 summarized the process of finding confusing category $c^*$ and generating the eval set. For simplicity, we call this new evaluation set as ImageNet with Adversarial Text Regions (ImageNet-Atr).
Fig~\ref{fig-imagenet-atr-example} shows a few examples of the ImageNet-Atr.

\begin{algorithm}[h!]
\SetKwFunction{BicubicInterp}{BicubicInterp}
\SetKwFunction{GradientDescent}{GradientDescent}
\SetKwInOut{Input}{Input}
\SetKwInOut{Output}{Output}
\BlankLine
\Input{50,000 images from ImageNet-1K evaluation set.}
\Output{A new eval set with 50,000 images, each including a spotted word on the image.}
\BlankLine
First sample 1\% of the ImageNet eval set. \\
    Use open-sourced CLIP model to evaluate the 1\% of data and calculate the confusion matrix $C$. \\
    \For{each class $c$ in $[1, 1000]$}{
        Find its most confusing class $c*$ using eq.\eqref{eq-most-confusing-category}
    }
\For{each image in the ImageNet eval set}{
    Given image's category $c$ and its most confusion category $c^*$, obtain the  word $w_{c^*}$ corresponding to  $c^*$ \\
    Add the word $w_{c^*}$ to the image at a random position.
}    
\caption{Generate the ImageNet-Atr Eval Set}
\end{algorithm}

\begin{figure*}[t!]
  \centering
  \includegraphics[height=6cm]{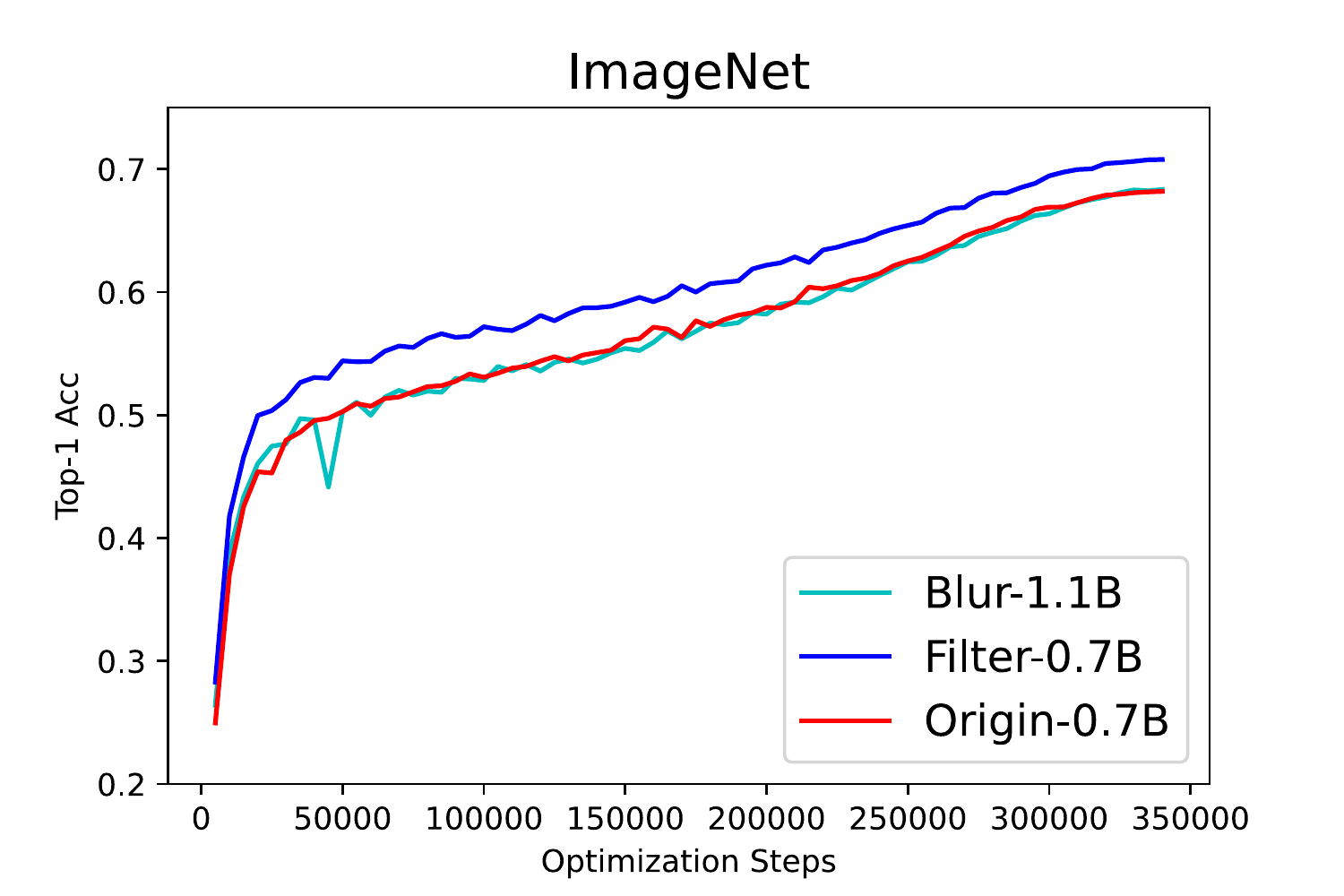}%
  \includegraphics[height=6cm]{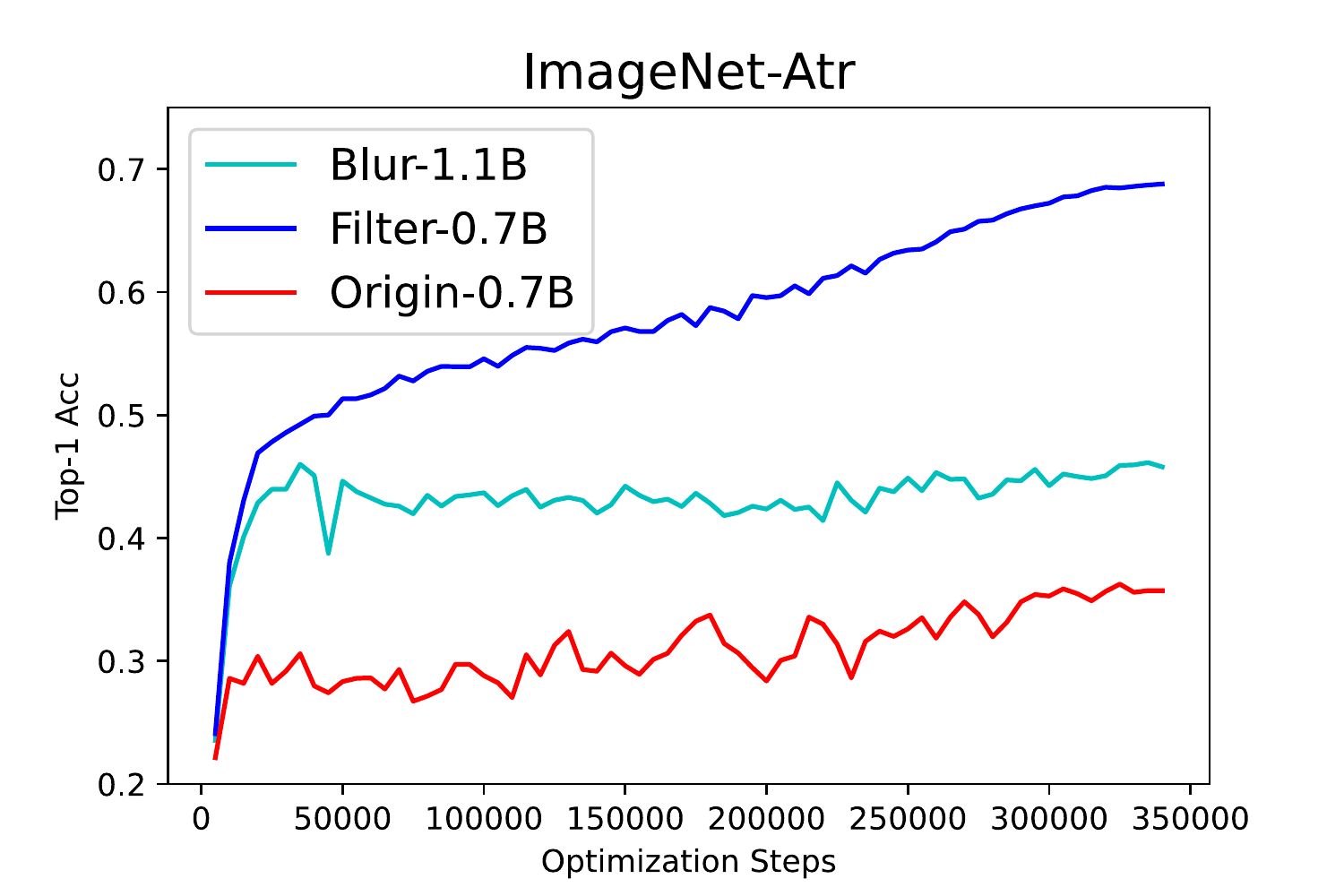}
  \caption{The Top-1 accuracy on ImageNet (left) and ImageNet-Atr (right) during the optimization.}
 \label{fig-imagenet-atr-opt}
\end{figure*}

\subsection{Evaluation Results}

Figure~\ref{fig-imagenet-atr-opt} shows the optimization process on ImageNet and ImageNet-Atr. The model from Filter-0.7B is trained to ignore the text regions, and it gets the highest accuracy on both ImageNet and ImageNet-Atr. In contrast, the model from Origin-0.7B gets reasonable accuracy on ImageNet, but much worse on ImageNet-Atr. This is because the model tends to focus more on text regions, so it gets confused when the text does not match the image semantics. At last, the accuracy of trained from Blur-1.1B is similar to Origin-0.7B on ImageNet, but becomes better on ImageNet-Atr. This is because the model was forced to not look at the image regions. However, its accuracy on ImageNet-Atr is still significantly worse than Filter-0.7B. These results suggest the simple filtering strategy will lead to the highest ImageNet accuracy and the most robust against typographic attacks. 

\begin{table}[h!]
  \caption{Comparing zero-shot classification accuracy. 
  } 
  \centering
  \begin{tabular}{l|c|cc}
    \toprule
   Model  & ImageNet & ImageNet-Atr \\
    \midrule
    OpenAI CLIP B-16          & 68.35\% & 31.65\% \\
    LAION CLIP B-16 &  66.99\%  &29.35\% \\
    Our CLIP (Origin-1.1B)  &68.66\% &35.73\% \\
    Our CLIP (Origin-0.7B)  &68.18\% &35.72\% \\
    Our CLIP (Blur-1.1B)  &68.34\% &45.78\% \\
    Our CLIP (Filter-0.7B) &\textbf{70.77\%} &\textbf{68.78\%} \\    
    \bottomrule
  \end{tabular}
\end{table}

\section{Discussion}

This paper suggests an easy-to-use method to improve CLIP training by filtering images with text regions. Despite its simpleness, the resulting model improves the accuracy of ImageNet from 68.66\% to 70.77\%, as well as better performances in other retrieval and linear probing benchmarks. In addition, this model is much more robust against typographic attacks. On our newly collected ImageNet with Adversarial Text Regions (ImageNet-Atr), this model's accuracy is 68.78\%, comparable with the accuracy on ImageNet. In contrast, the baseline CLIP model's accuracy on ImageNet-Atr is only 35.73\%. 

One potential limitation of the proposed approach is that it will overlook text regions that are correlated with image semantics, such as the title words on a picture of books. However, we'd suggest separating visual semantic and text region understanding and employing a separate OCR model for text region understanding for the latter task. 

Another potential limitation is that the proposed approach will reduce the training data size. Especially when the training model grows with a bigger capacity, this filtering approach may not become as significant as for smaller neural networks. To show this, we study the performance using CLIP-L/14 with 400M parameters and compare the performance in Table~\ref{table-compare-l14}. We can see that L/14 trained from Filtered-0.7B still gets the best accuracy, but its improvement on ImageNet (0.3\%) becomes smaller than the gain of the B/16 model (2.0\%). We will leave the other larger models for future study. 
 
\begin{table}[h!]
  \caption{\bf Comparing zero-shot classification accuracy of large models (CLIP L/14).} 
  \centering
  \begin{tabular}{l|cc}
    \toprule
   Model  & ImageNet & ImageNet-Atr \\
    \midrule
    L14 Origin-1.1B  &75.99\% &41.50\% \\
    L14 Origin-0.7B   &75.13\% &40.94\% \\
    L14 Filtered-0.7B  &\textbf{76.29\%} & \textbf{74.55\%} \\
    \bottomrule
  \end{tabular}
  \label{table-compare-l14}
\end{table}

\section*{Acknowledgement}
We want to thank  our colleagues Albin Madappally Jose, Futang Peng, Angie Wang, Jonathon Shlens, Mark Lee, Ruoming Pang, and Yang Zhao. We also appreciate the help from Zhe Gan, Hao Shen and Chong Wang who read the drat of this work and gave many valuable suggestions.

\section{Conclusion}
This paper considers the problems of improving CLIP training with a fixed optimization budget and proposes to enlarge the training set together and filter out data with text regions. This simple approach helps to boost the top-1 accuracy on ImageNet from 68.18\% to 70.77\%, together with other improvements on retrieval and linear probing tasks.

This paper also builds a new evaluation set named ImageNet-Atr, which can help us to benchmark the robustness against the typographic attack. We benchmark the open-sourced CLIP model and our internally trained CLIP models on this new eval dataset. Almost all the model's top-1 accuracy measures are lower than 50\%, except the model from Filter-0.7B gets a high accuracy of 68.78\%.

\bibliographystyle{ieee_fullname}

\bibliography{bibs/learning,bibs/clip,bibs/llm,bibs/data_selection,bibs/transformer,bibs/ocr}

\begin{thebibliography}{10}\itemsep=-1pt

\bibitem{ocr-craft-cvpr2019}
Youngmin Baek, Bado Lee, Dongyoon Han, Sangdoo Yun, and Hwalsuk Lee.
\newblock Character region awareness for text detection.
\newblock In {\em CVPR}, pages 9365--9374, 2019.

\bibitem{bottou2010-sgd}
L{\'e}on Bottou.
\newblock Large-scale machine learning with stochastic gradient descent.
\newblock In {\em Proceedings of COMPSTAT'2010: 19th International Conference
  on Computational StatisticsParis France, August 22-27, 2010 Keynote, Invited
  and Contributed Papers}, pages 177--186. Springer, 2010.

\bibitem{changpinyo2021cc12m}
Soravit Changpinyo, Piyush Sharma, Nan Ding, and Radu Soricut.
\newblock {Conceptual 12M}: Pushing web-scale image-text pre-training to
  recognize long-tail visual concepts.
\newblock In {\em Proceedings of the IEEE/CVF Conference on Computer Vision and
  Pattern Recognition}, pages 3558--3568, 2021.

\bibitem{chen2023stair}
Chen Chen, Bowen Zhang, Liangliang Cao, Jiguang Shen, Tom Gunter,
  Albin~Madappally Jose, Alexander Toshev, Jonathon Shlens, Ruoming Pang, and
  Yinfei Yang.
\newblock Stair: Learning sparse text and image representation in grounded
  tokens, 2023.

\bibitem{chen2023symbolic}
Xiangning Chen, Chen Liang, Da Huang, Esteban Real, Kaiyuan Wang, Yao Liu, Hieu
  Pham, Xuanyi Dong, Thang Luong, Cho-Jui Hsieh, et~al.
\newblock Symbolic discovery of optimization algorithms.
\newblock {\em arXiv preprint arXiv:2302.06675}, 2023.

\bibitem{kenton2019bert}
Jacob Devlin, Ming-Wei Chang, Kenton Lee, and Kristina Toutanova.
\newblock Bert: Pre-training of deep bidirectional transformers for language
  understanding.
\newblock In {\em Proceedings of NAACL-HLT}, pages 4171--4186, 2019.

\bibitem{vit2020}
Alexey Dosovitskiy, Lucas Beyer, Alexander Kolesnikov, Dirk Weissenborn,
  Xiaohua Zhai, Thomas Unterthiner, Mostafa Dehghani, Matthias Minderer, Georg
  Heigold, Sylvain Gelly, et~al.
\newblock An image is worth 16x16 words: Transformers for image recognition at
  scale.
\newblock {\em ICLR}, 2021.

\bibitem{goh2021multimodal}
Gabriel Goh, Nick~Cammarata †, Chelsea~Voss †, Shan Carter, Michael Petrov,
  Ludwig Schubert, Alec Radford, and Chris Olah.
\newblock Multimodal neurons in artificial neural networks.
\newblock {\em Distill}, 2021.
\newblock https://distill.pub/2021/multimodal-neurons.

\bibitem{Chinchilla2022}
Jordan Hoffmann, Sebastian Borgeaud, Arthur Mensch, Elena Buchatskaya, Trevor
  Cai, Eliza Rutherford, Diego de~Las Casas, Lisa~Anne Hendricks, Johannes
  Welbl, Aidan Clark, et~al.
\newblock Training compute-optimal large language models.
\newblock {\em arXiv:2203.15556}, 2022.

\bibitem{jia2021scaling}
Chao Jia, Yinfei Yang, Ye Xia, Yi-Ting Chen, Zarana Parekh, Hieu Pham, Quoc Le,
  Yun-Hsuan Sung, Zhen Li, and Tom Duerig.
\newblock Scaling up visual and vision-language representation learning with
  noisy text supervision.
\newblock In {\em International Conference on Machine Learning}, pages
  4904--4916. PMLR, 2021.

\bibitem{kil2022prestu}
Jihyung Kil, Soravit Changpinyo, Xi Chen, Hexiang Hu, Sebastian Goodman,
  Wei-Lun Chao, and Radu Soricut.
\newblock Prestu: Pre-training for scene-text understanding.
\newblock {\em arXiv:2209.05534}, 2022.

\bibitem{li2021trocr}
Minghao Li, Tengchao Lv, Jingye Chen, Lei Cui, Yijuan Lu, Dinei Florencio, Cha
  Zhang, Zhoujun Li, and Furu Wei.
\newblock Trocr: Transformer-based optical character recognition with
  pre-trained models.
\newblock {\em arXiv preprint arXiv:2109.10282}, 2021.

\bibitem{reclip2023}
Runze Li, Dahun Kim, Bir Bhanu, and Weicheng Kuo.
\newblock Reclip: Resource-efficient clip by training with small images, 2023.

\bibitem{li2022scaling}
Yanghao Li, Haoqi Fan, Ronghang Hu, Christoph Feichtenhofer, and Kaiming He.
\newblock Scaling language-image pre-training via masking.
\newblock {\em arXiv:2212.00794}, 2022.

\bibitem{flip2023}
Yanghao Li, Haoqi Fan, Ronghang Hu, Christoph Feichtenhofer, and Kaiming He.
\newblock Scaling language-image pre-training via masking, 2023.

\bibitem{declip2021}
Yangguang Li, Feng Liang, Lichen Zhao, Yufeng Cui, Wanli Ouyang, Jing Shao,
  Fengwei Yu, and Junjie Yan.
\newblock Supervision exists everywhere: A data efficient contrastive
  language-image pre-training paradigm.
\newblock {\em arXiv:2110.05208}, 2021.

\bibitem{ocr-binary-detection-pami22}
Minghui Liao, Zhisheng Zou, Zhaoyi Wan, Cong Yao, and Xiang Bai.
\newblock Real-time scene text detection with differentiable binarization and
  adaptive scale fusion.
\newblock {\em IEEE Transactions on Pattern Analysis and Machine Intelligence},
  45(1):919--931, 2022.

\bibitem{ocr-gooogle-cvpr2022}
Shangbang Long, Siyang Qin, Dmitry Panteleev, Alessandro Bissacco, Yasuhisa
  Fujii, and Michalis Raptis.
\newblock Towards end-to-end unified scene text detection and layout analysis.
\newblock In {\em CVPR}, pages 1049--1059, 2022.

\bibitem{lu2022unsupervised}
Zhiyun Lu, Yongqiang Wang, Yu Zhang, Wei Han, Zhehuai Chen, and Parisa Haghani.
\newblock Unsupervised data selection via discrete speech representation for
  asr.
\newblock {\em arXiv preprint arXiv:2204.01981}, 2022.

\bibitem{laion2b}
OpenCLIP.
\newblock Reaching 80\% zero-shot accuracy with openclip: Vit-g/14 trained on
  laion-2b.
\newblock https://laion.ai/blog/giant-openclip/, 2023.

\bibitem{pham2021combined}
Hieu Pham, Zihang Dai, Golnaz Ghiasi, Hanxiao Liu, Adams~Wei Yu, Minh-Thang
  Luong, Mingxing Tan, and Quoc~V Le.
\newblock Combined scaling for zero-shot transfer learning.
\newblock {\em arXiv:2111.10050}, 2021.

\bibitem{radenovic2023filtering}
Filip Radenovic, Abhimanyu Dubey, Abhishek Kadian, Todor Mihaylov, Simon
  Vandenhende, Yash Patel, Yi Wen, Vignesh Ramanathan, and Dhruv Mahajan.
\newblock Filtering, distillation, and hard negatives for vision-language
  pre-training.
\newblock {\em arXiv preprint arXiv:2301.02280}, 2023.

\bibitem{clip}
Alec Radford, Jong~Wook Kim, Chris Hallacy, Aditya Ramesh, Gabriel Goh,
  Sandhini Agarwal, Girish Sastry, Amanda Askell, Pamela Mishkin, Jack Clark,
  et~al.
\newblock Learning transferable visual models from natural language
  supervision.
\newblock In {\em International conference on machine learning}, pages
  8748--8763, 2021.

\bibitem{Gopher2022}
Jack~W. Rae, Sebastian Borgeaud, Trevor Cai, Katie Millican, Jordan Hoffmann,
  Francis Song, John Aslanides, Sarah Henderson, Roman Ring, Susannah Young,
  Eliza Rutherford, Tom Hennigan, Jacob Menick, Albin Cassirer, Richard Powell,
  George van~den Driessche, Lisa~Anne Hendricks, Maribeth Rauh, Po-Sen Huang,
  Amelia Glaese, Johannes Welbl, Sumanth Dathathri, Saffron Huang, Jonathan
  Uesato, John Mellor, Irina Higgins, Antonia Creswell, Nat McAleese, Amy Wu,
  Erich Elsen, Siddhant Jayakumar, Elena Buchatskaya, David Budden, Esme
  Sutherland, Karen Simonyan, Michela Paganini, Laurent Sifre, Lena Martens,
  Xiang~Lorraine Li, Adhiguna Kuncoro, Aida Nematzadeh, Elena Gribovskaya,
  Domenic Donato, Angeliki Lazaridou, Arthur Mensch, Jean-Baptiste Lespiau,
  Maria Tsimpoukelli, Nikolai Grigorev, Doug Fritz, Thibault Sottiaux, Mantas
  Pajarskas, Toby Pohlen, Zhitao Gong, Daniel Toyama, Cyprien de
  Masson~d'Autume, Yujia Li, Tayfun Terzi, Vladimir Mikulik, Igor Babuschkin,
  Aidan Clark, Diego de Las~Casas, Aurelia Guy, Chris Jones, James Bradbury,
  Matthew Johnson, Blake Hechtman, Laura Weidinger, Iason Gabriel, William
  Isaac, Ed Lockhart, Simon Osindero, Laura Rimell, Chris Dyer, Oriol Vinyals,
  Kareem Ayoub, Jeff Stanway, Lorrayne Bennett, Demis Hassabis, Koray
  Kavukcuoglu, and Geoffrey Irving.
\newblock Scaling language models: Methods, analysis \& insights from training
  gopher, 2022.

\bibitem{sharma-etal-2018-conceptual}
Piyush Sharma, Nan Ding, Sebastian Goodman, and Radu Soricut.
\newblock Conceptual captions: A cleaned, hypernymed, image alt-text dataset
  for automatic image captioning.
\newblock In {\em Proceedings of the 56th Annual Meeting of the Association for
  Computational Linguistics (Volume 1: Long Papers)}, pages 2556--2565, 2018.

\bibitem{song2022vision}
Sibo Song, Jianqiang Wan, Zhibo Yang, Jun Tang, Wenqing Cheng, Xiang Bai, and
  Cong Yao.
\newblock Vision-language pre-training for boosting scene text detectors.
\newblock In {\em CVPR}, pages 15681--15691, 2022.

\bibitem{sorscher2022beyond}
Ben Sorscher, Robert Geirhos, Shashank Shekhar, Surya Ganguli, and Ari Morcos.
\newblock Beyond neural scaling laws: beating power law scaling via data
  pruning.
\newblock {\em Advances in Neural Information Processing Systems},
  35:19523--19536, 2022.

\bibitem{transformer}
Ashish Vaswani, Noam Shazeer, Niki Parmar, Jakob Uszkoreit, Llion Jones,
  Aidan~N Gomez, {\L}ukasz Kaiser, and Illia Polosukhin.
\newblock Attention is all you need.
\newblock {\em Advances in neural information processing systems}, 30, 2017.

\bibitem{xiao2022exploiting}
Taihong Xiao, Zirui Wang, Liangliang Cao, Jiahui Yu, Shengyang Dai, and
  Ming-Hsuan Yang.
\newblock Exploiting category names for few-shot classification with
  vision-language models.
\newblock {\em arXiv:2211.16594}, 2022.

\bibitem{xue2022languagematters}
Chuhui Xue, Wenqing Zhang, Yu Hao, Shijian Lu, Philip~HS Torr, and Song Bai.
\newblock Language matters: A weakly supervised vision-language pre-training
  approach for scene text detection and spotting.
\newblock In {\em ECCV}, pages 284--302, 2022.

\bibitem{yang2021tap}
Zhengyuan Yang, Yijuan Lu, Jianfeng Wang, Xi Yin, Dinei Florencio, Lijuan Wang,
  Cha Zhang, Lei Zhang, and Jiebo Luo.
\newblock Tap: Text-aware pre-training for text-vqa and text-caption.
\newblock In {\em CVPR}, pages 8751--8761, 2021.

\bibitem{yao2021filip}
Lewei Yao, Runhui Huang, Lu Hou, Guansong Lu, Minzhe Niu, Hang Xu, Xiaodan
  Liang, Zhenguo Li, Xin Jiang, and Chunjing Xu.
\newblock Filip: fine-grained interactive language-image pre-training.
\newblock {\em arXiv:2111.07783}, 2021.

\bibitem{yu2022coca}
Jiahui Yu, Zirui Wang, Vijay Vasudevan, Legg Yeung, Mojtaba Seyedhosseini, and
  Yonghui Wu.
\newblock Coca: Contrastive captioners are image-text foundation models.
\newblock {\em arXiv:2205.01917}, 2022.

\bibitem{zhai2022scaling}
Xiaohua Zhai, Alexander Kolesnikov, Neil Houlsby, and Lucas Beyer.
\newblock Scaling vision transformers.
\newblock In {\em Proceedings of the IEEE/CVF Conference on Computer Vision and
  Pattern Recognition}, pages 12104--12113, 2022.

\bibitem{zhai2023sigmoid}
Xiaohua Zhai, Basil Mustafa, Alexander Kolesnikov, and Lucas Beyer.
\newblock Sigmoid loss for language image pre-training.
\newblock {\em arXiv preprint arXiv:2303.15343}, 2023.

\bibitem{xCLIP2022}
Jinghao Zhou, Li Dong, Zhe Gan, Lijuan Wang, and Furu Wei.
\newblock Non-contrastive learning meets language-image pre-training.
\newblock {\em arXiv:2210.09304}, 2022.

\end{thebibliography}

\end{document}